# Are LLMs Good Text Diacritizers? An Arabic and Yoruba Case Study

**Hawau Olamide Toyin**[1], **Samar Mohamed Magdy**[2], **Hanan Aldarmaki**[1]
[1]Mohamed Bin Zayed University of Artificial Intelligence, UAE
[2]University of British Columbia, Canada
{hawau.toyin, hanan.aldarmaki}@mbzuai.ac.ae

**Abstract**

We investigate the effectiveness of large language models (LLMs) for text diacritization in two typologically distinct languages: Arabic and Yoruba. To enable a rigorous evaluation, we introduce a novel multilingual dataset `MultiDiac`, with diverse samples that capture a range of diacritic ambiguities. We evaluate $12$ LLMs varying in size, accessibility, and language coverage, and benchmark them against $4$ specialized diacritization models. Additionally, we fine-tune four small open-source models using LoRA for Yoruba. Our results show that many off-the-shelf LLMs outperform specialized diacritization models, but smaller models suffer from hallucinations. We find that fine-tuning on a small dataset can help improve diacritization performance and reduce hallucinations for Yoruba.

## 1. Introduction

Arabic, a Semitic language, and Yoruba, a Niger-Congo language, in their written scripts rely heavily on diacritics for disambiguation, though their functions differ considerably. In Arabic, diacritics primarily represent short vowels and consonant doubling, and are typically omitted in everyday writing. In contrast, Yoruba employs diacritics, specifically tone marks and vowel diacritics, to encode lexical tone and vowel quality, both of which are essential for distinguishing meaning in this tonal language (see Table 1). The omission of diacritics in both languages leads to significant ambiguity, especially for language learning, underscoring the importance of automatic diacritization.

Typically, specialized models are trained for automatic text diacritization, requiring dedicated data and training efforts (e.g., Orife, 2018; Shatnawi et al., 2024; Skiredj and Berrada, 2024). *This paper explores whether LLMs can perform diacritization effectively*. Given their extensive text exposure, existing publicly available diacritized corpora may overlap with LLM pre-training data, potentially inflating evaluation results and obscuring true model generalization capabilities. To effectively evaluate LLMs for diacritization, it is crucial to construct datasets that provide *novel* and diverse samples with systematically varying linguistic ambiguity beyond what the models have encountered during pre-training. In this vein, we introduce `MultiDiac`, a carefully curated multilingual test set designed to rigorously evaluate LLM-based diacritization in Arabic and Yoruba, featuring novel and diverse samples to minimize overlap with existing LLM pre-training data. Furthermore, we present the first large-scale evaluation of 12 LLMs on diacritization tasks across two typologically distinct languages, benchmarking against specialized models and providing a comprehensive analysis of general-purpose LLM capabilities in this under-explored task. Finally, we demonstrate that fine-tuning small, open-source LLMs via LoRA substantially improves performance and reduces hallucination for the low-resource Yoruba language.

| Without Diacritics | Possible Interpretations |
|---|---|
| `Aje` wo ile re | ① A prayer (Wealth entered your house)<br>② A curse (A witch entered your house) |

Table 1: In this Yoruba text, ambiguities stem from not diacritizing the word `Aje`, with diacritics: `Ajé` (wealth) or `Àjẹ́` (witch).

## 2. Related Work

**Arabic Diacritization.** Most of the existing work on diacritization build resources and specialized models that take undiacritized text as input and predict the corresponding diacritics (e.g., Darwish et al., 2021, 2017; Alasmary et al., 2024). While most of the resources focus on Modern Standard Arabic (Alasmary et al., 2024), few target Arabic dialects (Talafha et al., 2025). In addition, some recent efforts explored using the speech modality to complement text resources for Arabic diacritization Aldarmaki and Ghannam (2023); **?**); Shatnawi et al. (2024).In terms of extrinsic evaluation, some studies have shown that diacritization improves Arabic machine translation (Fadel et al., 2019; Kadaoui et al., 2023).

**Yoruba Diacritization.** Research on Yorùbá text diacritization goes from rule- and syllable-based (Adegbola and Odilinye, 2012; Asahiah et al., 2017) methods to modern neural approaches (Ola-wole et al., 2024; Orife et al., 2020). Initially, researchers used diacritized Bible verses in the Yoruba lanuguage as a training resource. Adelani et al. investigated the effect of data domain and text diacritization on Machine Translation. Olawole et al. recently introduced a multi-domain diacritized Yoruba text for research.

**LLMs for Different Tasks.** LLMs have demonstrated remarkable capabilities across a wide range of natural language processing tasks, primarily in text understanding and generation. Previous studies have shown that LLMs achieve state-of-the-art performance in tasks such as machine translation (Brown et al., 2020; Chen et al., 2021), summarization (Goyal et al., 2022), question answering (Yue, 2025), and paraphrase generation (Yadav et al., 2024). Their ability to capture rich contextual representations has also made them effective in zero-shot and few-shot learning settings, where they can generalize to new tasks with minimal supervision (Brown et al., 2020). Beyond text-only tasks, LLMs have been applied to code generation, reasoning, and multimodal tasks (OpenAI et al., 2024). However, despite their widespread use in text processing tasks, their application to fine-grained orthographic tasks such as text diacritization, which requires precise surface-level predictions grounded in linguistic context, remains underexplored.

## 3. The MultiDiac Dataset

We constructed a new dataset that comprises fully diacritized meaningful sentences formed manually by annotators around a specific word. The curation process involves mainly two steps: (1) base word identification, and (2) manual sentence formation to contextualize the base word. We define a base word as a word that, in its undiacritized form, (*with accent and diacritic marks stripped)* and without surrounding content, could have multiple pronunciations and meanings. A sample of a base word is shown for both languages in Figure 1 with their possible diacritized forms.

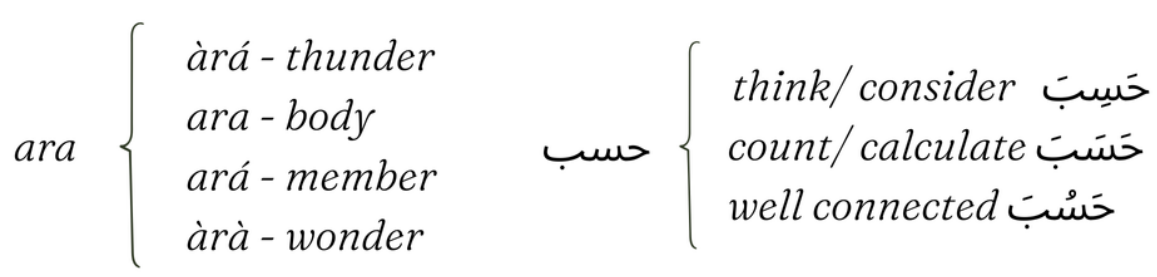

Figure 1: Sample of base word with multiple diacritization variants in Yoruba (left) and Arabic (right).

**Evaluation Set.** The `Yoruba` sentences were formed in collaboration with three native speakers who hold higher academic degrees in the language. Approximately 71 base words with multiple valid diacritized forms (see Figure 1 for an example) were first identified. For each base word, a contextually rich sentence, was constructed from 1 to 2 diacritized forms. The resulting test set of 100 fully diacritized sentences was manually verified by an independent native speaker and the lead annotator for accuracy.

The `Arabic` portion of the dataset was developed in collaboration with a native Arabic speaker and a proficient L2 speaker. A total of 42 base words were initially identified. For each word, $2-3$ diacritized variants were identified, and contextually diverse sentences were crafted to capture distinct semantic interpretations. All sentences were subsequently reviewed and validated by the native speaker for linguistic accuracy, grammaticality, and naturalness. The finalized gold-standard test set comprises 106 fully diacritized Arabic sentences.

**Training Set.** In addition to the test set for both languages, we developed a small Training set for Yoruba as it's a low-resource language and preliminary evaluation showed poor performance compared to Arabic. The train set comprises 164 base words and 603 sentences constructed from between $2-4$ diacritized forms of each base word. We quantify the diversity of our proposed train set by measuring the n-gram overlap ($1 \leq n \leq 3$) between our train set and existing Yoruba diacritized text resources[1] used to train specialized diacritization models using word-level tokenization. The 1-gram overlap was 79% indicating shared vocabulary, while the 2- and 3-gram overlaps dropped to 47% and 13% respectively, indicating minimal phrase-level similarity. We focused solely on a test set for `Arabic`, given its relatively higher resource availability compared to Yoruba (Joshi et al., 2020) and its better performance.

For both languages, the dataset contains sentences that are, *on average, seven words long*. We present one sample from each language in the dataset in Table 2. The data is publicly available.[2]

| **Lang** | **Base** | **Base_definition** | **Text** |
|---|---|---|---|
| Arabic | حَسِبَ | think/consider | أَمْ حَسِبَ الَّذِينَ كَفَرُوا أَنَّ لَهُمْ مَخْرَجًا |
| Yoruba | Owò | business | Gbogbo ẹni ti o ṣe owò ni yíò jèrè rẹ̀ |

Table 2: Sample data from our dataset.

## 4. Experimental Setup

### 4.1. Prompt Variety

The prompts used to instruct an LLM are important to obtain the desired outcome for a particular task

[1] https://github.com/ajesujoba/YAD
[2] https://huggingface.co/datasets/herwoww/MultiDiac

| Prompt | Template |
|---|---|
| P1 | Add diacritics and accent marks to this text: {text} Return the sentence only! |
| P2 | Add diacritics and accent marks to this {lang} sentence: {text} Return the sentence only! |
| P3 | Add diacritics and accent marks to this {lang} sentence, the word '{base}' in the sentence means '{base_definition}' {text} Return the sentence only! Don't add any explanations!. |

Table 3: Prompt templates used for the LLM diacritization experiments. Placeholders ({text}, {base}, {base_definition}, {lang} ) are filled dynamically with example content during inference (see Table 2).

(Kadaoui et al., 2023). Kadaoui et al. reported that English prompts yielded better results than Arabic prompts for their Arabic machine translation evaluation of LLMs. To determine the extent of a models performance variance, we vary the prompt in 3 ways: (P1) language-agnostic, (P2) language-specific , and (P3) pseudo translation-assisted, as shown in Table 3.

## 4.2. Experiments

To comprehensively assess the models' capabilities, we selected LLMs that vary along three key dimensions: *model size*, *availability*, and *language coverage*. Specifically, we include both small and large models to examine the impact of model capacity, as well as a mix of open-source and closed-source models to account for accessibility and architectural diversity. Furthermore, we evaluate models with different linguistic pre-training scopes, including Arabic-specific models and multilingual models, to analyze how language specialization influences diacritization performance. We report results averaged over 3 runs.

**LLMs.** We evaluate a total of $12$ LLMs. Subsequently, in this paper, we refer to models with a total number of parameters less than 15B as `SmallLMs`, and those with a total number of parameters greater than or equal to 15B as `LargeLMs`. Of the $7$ `SmallLMs`, three are Arabic-centric LLMs developed to improve Arabic text generation: *Allam-7B* (Bari et al., 2024), *Jais-13B* (Sengupta et al., 2023), and *Fanar-9B* (Team et al., 2025). With testing these models, we're able to evaluate the performance of language centric LLMs. The other `SmallLMs` are: Qwen2.5, Gemma-7B, LLama-1B and Phi-4 (Qwen et al., 2025; Team et al., 2024; Grattafiori et al., 2024; Abdin et al., 2024). All `SmallLMs` were evaluated through their official HuggingFace repo. All `LargeLMs` (*Claude3.5* (Anthropic, 2024),*IBM granite*[3], *Deepseek-R1* (DeepSeek-AI et al., 2025),*Command-a* (Cohere., 2024),*ChatGPT4o* (OpenAI et al., 2024)), were run from the `LM arena`[4] except *ChatGPT4o*, which was run from the official chat interface.

**Baselines.** As baselines, we evaluate existing state-of-the-art `specialized` models for text diacritization and compare them against the LLMs evaluated in this work. For Arabic, we compare against *CATT*[5] (Alasmary et al., 2024) and *Shakkeelha*[6] (Fadel et al., 2019). *Oyo-T5*[7] and *byT5*[8] from Olawole et al. serve as baselines for Yoruba. All models are publicly available for use in research.

**LoRA Fine-tuning.** To adapt open-source LLMs for the diacritization task described previously, we employed parameter-efficient fine-tuning, namely Low-Rank Adaptation (LoRA) (Hu et al., 2021), using our Yoruba training subset with prompt `P2`. Our LoRA configuration targeted key projection and feedforward layers in the transformer architecture, with a rank of $16$, a scaling factor of $32$, and a dropout rate of 0.05. All experiments were completed using a single 40GB A100 GPU.

## 4.3. Evaluation

**Diacritization.** We report the character error rate (CER), which is proportional to the diacritic error rate (DER) when the underlying characters are unchanged (*which is the case with specialized models*). We utilize CER to extend this metric to generative models(LLMs), which may include character errors, and for Yoruba where diacritization means new latin characters. The CER measures the edit distance between the predicted and reference sequences at the character level.

**Hallucination.** In addition, we assess the extent of **hallucinations in the LLM outputs**, where models generate output that is not grounded in the input (Zhang et al., 2023). Specifically, we quantify hallucinations by computing the WER between the predicted text (*with all diacritics removed*) and the reference (*with all diacritics removed*), capturing alterations in the underlying text.

[3] https://www.ibm.com/granite
[4] https://lmarena.ai/
[5] https://github.com/abjadai/catt
[6] https://github.com/AliOsm/shakkelha
[7] https://huggingface.co/Davlan/omowe-t5-small-diacritizer-all-und-full
[8] Davlan/byt5-small-diacritizer-menyo

## 5. Results and Discussion

**Are LLMs good text diacritizers?** *Yes,* while `SmallLMs` generally have poor CER (20-60%), `LargeLMs` are better on average, with CER ranging from $2-30\%$ for both languages. For Arabic, the maximum CER at $5.81\%$ across all models tested, compared to over $10\%$ CER for specialized models. This is not the case for Yoruba, with a minimum CER at $13.65\%$ across all models tested, but they are still far better than specialized models, even the `SmallLMs`. The deviation in actual diacritization errors across $3$ runs is not significant in most models. **Is it better to use a language-centric LLMs?** *Yes,* given access to only small LLMs. Arabic-centric LLMs show better performance than all small LLMs we evaluated.

| | Model \ Lang. | CER↓ Yoruba | CER↓ Arabic | WER↓ Yoruba | WER↓ Arabic |
|---|---|---|---|---|---|
| Specialized | CATT | - | 10.3 | - | - |
| | Shakkelha | - | 11.5 | - | - |
| | Oyo-T5 | 52.8 | - | - | - |
| | byT5 | 46.5 | - | - | - |
| SmallLMs | Llama-1B | 47.3 $^{\pm 4.9}$ | 49.0 $^{\pm 2.6}$ | 43.5 $^{\pm 5.0}$ | 35.8 $^{\pm 11.5}$ |
| | Gemma-7B | 35.3 $^{\pm 0}$ | 45.4 $^{\pm 0}$ | 35.3 $^{\pm 0}$ | 28.5 $^{\pm 0}$ |
| | Qwen2.5 | 38.6 $^{\pm 1.1}$ | 27.5 $^{\pm 1.6}$ | 38.6 $^{\pm 1.1}$ | 11.5 $^{\pm 1.4}$ |
| | Phi-4 | 30.2 $^{\pm 0}$ | 14.2 $^{\pm 0}$ | 30.2 $^{\pm 0}$ | 9.7 $^{\pm 0}$ |
| | Allam-9B | - | 20.8 $^{\pm 0}$ | - | 20.5 $^{\pm 0}$ |
| | Fanar-9B | - | 7.5 $^{\pm 0}$ | - | 4.5 $^{\pm 0}$ |
| | Jais-13B | - | 61.9 $^{\pm 1.1}$ | - | 75.5 $^{\pm 1.4}$ |
| LargeLMs | Claude3.5 | 20.4 $^{\pm 1.3}$ | 5.8 $^{\pm 0.4}$ | 21.4 $^{\pm 0.3}$ | 3.4 $^{\pm 3.9}$ |
| | IBM granite | 26.1 $^{\pm 0.1}$ | 3.2 $^{\pm 0.02}$ | 25.9 $^{\pm 0.0}$ | 1.1 $^{\pm 0}$ |
| | Deepseek-R1 | 14.9 $^{\pm 0.2}$ | 2.6 $^{\pm 0.9}$ | 13.9 $^{\pm 3.0}$ | 2.5 $^{\pm 3.4}$ |
| | Command-a | 28.7 $^{\pm 0.5}$ | 5.7 $^{\pm 0.03}$ | 22.7 $^{\pm 2.0}$ | 1.1 $^{\pm 0}$ |
| | ChatGPT4o | 13.6 $^{\pm 0.7}$ | 2.0 $^{\pm 0.8}$ | 15.7 $^{\pm 3.4}$ | 1.2 $^{\pm 0.8}$ |

Table 4: Average evaluation results over 3 runs for LLMs using `P2` and results from `specialized` diacritization models for both languages.

**Is there a better prompt?** *Yes,* the aggregated trend in Figure 2 shows that richer prompts (`P2,P3`) tend to reduce hallucination (WER) and improve diacritization (CER), particularly for Arabic models, while Yoruba systems exhibit higher variability but similar qualitative gains. All models perform better when the language is specified in the prompt, except Qwen. For Qwen on Yorùbá, paired t-tests on `P1,P2` and `P2,P3` showed no statistically significant improvement, with a p-value of $\approx 0.2$ .

**The hallucinations.** For `Arabic`, the high WER ($> 50\%$) reported is mainly linked to the Jais model returning transliterations or translations of the input sentence instead of the diacritized Arabic sentences. In other cases, the models return modified forms of the original sentence in terms of sentence length. The smallest model (Llama-1B) also returns output in characters of other languages (Hindi, Russian) when `P1` is used. For `P2`, the occurrence of other languages was reduced; however, there were also translations. For `Yoruba`, a clear drop in hallucination WER was noted when the language was specified in the prompt (`P2,P3`) for most models (see Figure 2). For `P1`, most hallucinations stemmed from the models' misidentifying the language of the sentence. This leads to the use of incorrect accent marks from other languages (Italian, Spanish). We can significantly improve performance for Yoruba with Lora Fine-tuning, as show in Table 5.

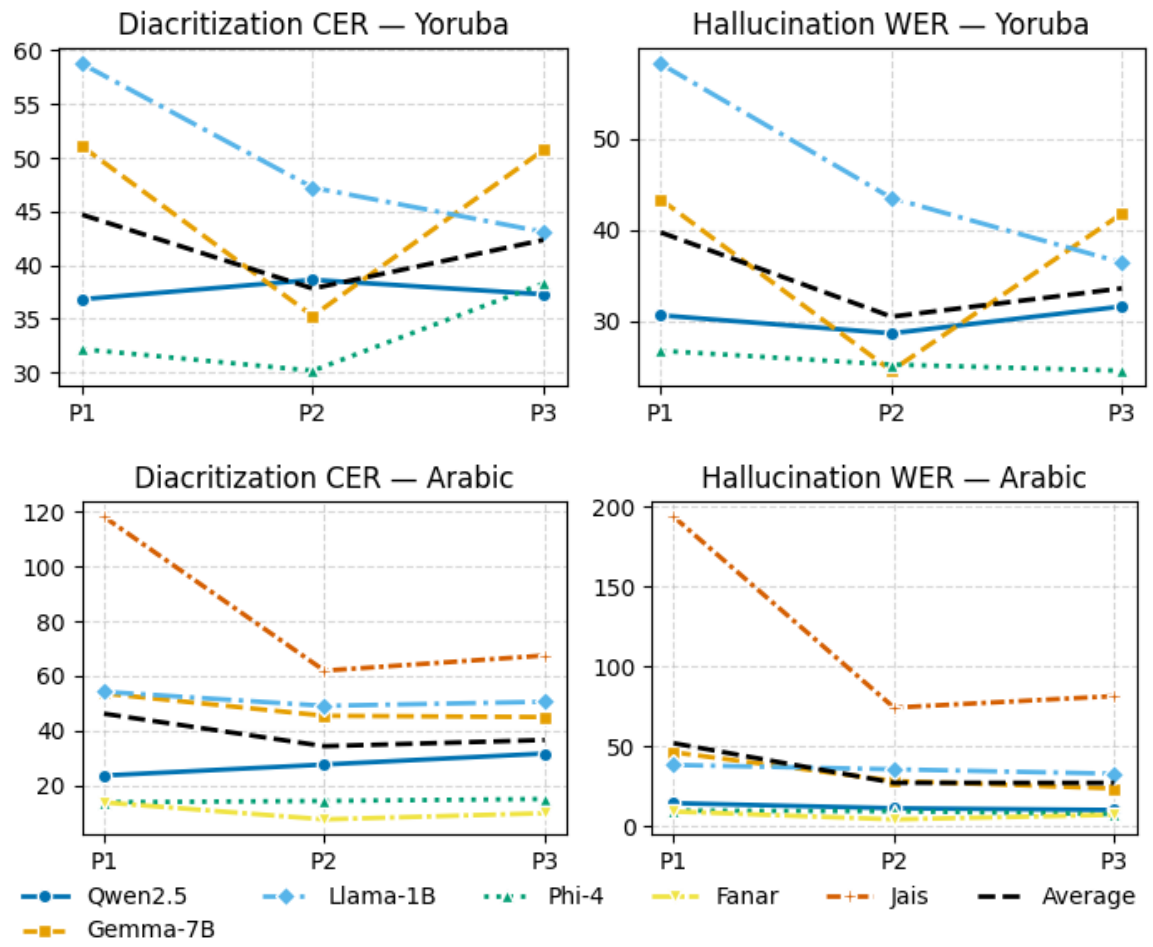

Figure 2: CER and WER models and languages. Each colored line represents a single model's mean performance across 3 runs, while the black dashed line is the average across all models. The x-axis corresponds to the three prompt formulations as defined in Section 4.1.

| | Original CER↓ | Original WER↓ | post-LoRA CER↓ | post-LoRA WER↓ |
|---|---|---|---|---|
| **Llama-1B** | 47.3 $^{\pm 4.9}$ | 43.5 $^{\pm 5.0}$ | 17.3 $^{\pm 0.7}$ | 12.5 $^{\pm 0.3}$ |
| **Gemma-7B** | 35.3 $^{\pm 0}$ | 35.3 $^{\pm 0}$ | 19.8 $^{\pm 0.3}$ | 7.3 $^{\pm 0.4}$ |
| **Phi-4** | 30.2 $^{\pm 0}$ | 30.2 $^{\pm 0}$ | 12.6 $^{\pm 12.6}$ | 2.7 $^{\pm 0.2}$ |
| **Qwen2.5** | 38.6 $^{\pm 1.1}$ | 38.6 $^{\pm 1.1}$ | 17.4 $^{\pm 0.6}$ | 3.6 $^{\pm 0.4}$ |

Table 5: Diacritization and hallucination results over 3 runs on the Yoruba dataset pre and post LoRA fine-tuning.

## 6. Conclusion

In this study, we investigated the capability of large language models as contextual text diacritizers through a focused case study on Arabic and Yoruba. We introduced a novel multilingual evaluation dataset specifically designed to capture diverse diacritic ambiguities in both languages. Our comprehensive evaluation spanned seven small and five large language models, encompassing both open-source and closed-source systems, and benchmarked their performance against existing state-of-the-art specialized diacritization models.

## 7. Ethical Considerations and Limitations

**Data reproducibility:** since the annotators randomly create sentences from the base words, the dataset construction is not exactly reproducible. We will provide the dataset for other researchers who wish to exactly reproduce our results.
**Variations in Models:** LLMs are trained with different datasets, and specialized diacritization models are trained on specialized diacritic restoration data sets. The models evaluated are not comparable one-to-one. Our results should not be taken to imply anything about the modelling methodology more generally, but rather about the current capabilities of *existing* LLMs and specialized models.
**Language Coverage:** While we evaluate two languages, there are other languages that use diacritics which were not included in this study.
**Compensation:** The language experts were fairly compensated for contributing to this work.

Praveen Krishnan, Punit Singh Koura, Puxin Xu, Qing He, Qingxiao Dong, Ragavan Srinivasan, Raj Ganapathy, Ramon Calderer, Ricardo Silveira Cabral, Robert Stojnic, Roberta Raileanu, Rohan Maheswari, Rohit Girdhar, Rohit Patel, Romain Sauvestre, Ronnie Polidoro, Roshan Sumbaly, Ross Taylor, Ruan Silva, Rui Hou, Rui Wang, Saghar Hosseini, Sahana Chennabasappa, Sanjay Singh, Sean Bell, Seohyun Sonia Kim, Sergey Edunov, Shaoliang Nie, Sharan Narang, Sharath Raparthy, Sheng Shen, Shengye Wan, Shruti Bhosale, Shun Zhang, Simon Vandenhende, Soumya Batra, Spencer Whitman, Sten Sootla, Stephane Collot, Suchin Gururangan, Sydney Borodinsky, Tamar Herman, Tara Fowler, Tarek Sheasha, Thomas Georgiou, Thomas Scialom, Tobias Speckbacher, Todor Mihaylov, Tong Xiao, Ujjwal Karn, Vedanuj Goswami, Vibhor Gupta, Vignesh Ramanathan, Viktor Kerkez, Vincent Gonguet, Virginie Do, Vish Vogeti, Vítor Albiero, Vladan Petrovic, Weiwei Chu, Wenhan Xiong, Wenyin Fu, Whitney Meers, Xavier Martinet, Xiaodong Wang, Xiaofang Wang, Xiaoqing Ellen Tan, Xide Xia, Xinfeng Xie, Xuchao Jia, Xuewei Wang, Yaelle Goldschlag, Yashesh Gaur, Yasmine Babaei, Yi Wen, Yiwen Song, Yuchen Zhang, Yue Li, Yuning Mao, Zacharie Delpierre Coudert, Zheng Yan, Zhengxing Chen, Zoe Papakipos, Aaditya Singh, Aayushi Srivastava, Abha Jain, Adam Kelsey, Adam Shajnfeld, Adithya Gangidi, Adolfo Victoria, Ahuva Goldstand, Ajay Menon, Ajay Sharma, Alex Boesenberg, Alexei Baevski, Allie Feinstein, Amanda Kallet, Amit Sangani, Amos Teo, Anam Yunus, Andrei Lupu, Andres Alvarado, Andrew Caples, Andrew Gu, Andrew Ho, Andrew Poulton, Andrew Ryan, Ankit Ramchandani, Annie Dong, Annie Franco, Anuj Goyal, Aparajita Saraf, Arkabandhu Chowdhury, Ashley Gabriel, Ashwin Bharambe, Assaf Eisenman, Azadeh Yazdan, Beau James, Ben Maurer, Benjamin Leonhardi, Bernie Huang, Beth Loyd, Beto De Paola, Bhargavi Paranjape, Bing Liu, Bo Wu, Boyu Ni, Braden Hancock, Bram Wasti, Brandon Spence, Brani Stojkovic, Brian Gamido, Britt Montalvo, Carl Parker, Carly Burton, Catalina Mejia, Ce Liu, Changhan Wang, Changkyu Kim, Chao Zhou, Chester Hu, Ching-Hsiang Chu, Chris Cai, Chris Tindal, Christoph Feichtenhofer, Cynthia Gao, Damon Civin, Dana Beaty, Daniel Kreymer, Daniel Li, David Adkins, David Xu, Davide Testuggine, Delia David, Devi Parikh, Diana Liskovich, Didem Foss, Dingkang Wang, Duc Le, Dustin Holland, Edward Dowling, Eissa Jamil, Elaine Montgomery, Eleonora Presani, Emily Hahn, Emily Wood, Eric-Tuan Le, Erik Brinkman, Esteban Arcaute, Evan Dunbar, Evan Smothers, Fei Sun, Felix Kreuk, Feng Tian, Filippos Kokkinos, Firat Ozgenel, Francesco Caggioni, Frank Kanayet, Frank Seide, Gabriela Medina Florez, Gabriella Schwarz, Gada Badeer, Georgia Swee, Gil Halpern, Grant Herman, Grigory Sizov, Guangyi, Zhang, Guna Lakshminarayanan, Hakan Inan, Hamid Shojanazeri, Han Zou, Hannah Wang, Hanwen Zha, Haroun Habeeb, Harrison Rudolph, Helen Suk, Henry Aspegren, Hunter Goldman, Hongyuan Zhan, Ibrahim Damlaj, Igor Molybog, Igor Tufanov, Ilias Leontiadis, Irina-Elena Veliche, Itai Gat, Jake Weissman, James Geboski, James Kohli, Janice Lam, Japhet Asher, Jean-Baptiste Gaya, Jeff Marcus, Jeff Tang, Jennifer Chan, Jenny Zhen, Jeremy Reizenstein, Jeremy Teboul, Jessica Zhong, Jian Jin, Jingyi Yang, Joe Cummings, Jon Carvill, Jon Shepard, Jonathan McPhie, Jonathan Torres, Josh Ginsburg, Junjie Wang, Kai Wu, Kam Hou U, Karan Saxena, Kartikay Khandelwal, Katayoun Zand, Kathy Matosich, Kaushik Veeraraghavan, Kelly Michelena, Keqian Li, Kiran Jagadeesh, Kun Huang, Kunal Chawla, Kyle Huang, Lailin Chen, Lakshya Garg, Lavender A, Leandro Silva, Lee Bell, Lei Zhang, Liangpeng Guo, Licheng Yu, Liron Moshkovich, Luca Wehrstedt, Madian Khabsa, Manav Avalani, Manish Bhatt, Martynas Mankus, Matan Hasson, Matthew Lennie, Matthias Reso, Maxim Groshev, Maxim Naumov, Maya Lathi, Meghan Keneally, Miao Liu, Michael L. Seltzer, Michal Valko, Michelle Restrepo, Mihir Patel, Mik Vyatskov, Mikayel Samvelyan, Mike Clark, Mike Macey, Mike Wang, Miquel Jubert Hermoso, Mo Metanat, Mohammad Rastegari, Munish Bansal, Nandhini Santhanam, Natascha Parks, Natasha White, Navyata Bawa, Nayan Singhal, Nick Egebo, Nicolas Usunier, Nikhil Mehta, Nikolay Pavlovich Laptev, Ning Dong, Norman Cheng, Oleg Chernoguz, Olivia Hart, Omkar Salpekar, Ozlem Kalinli, Parkin Kent, Parth Parekh, Paul Saab, Pavan Balaji, Pedro Rittner, Philip Bontrager, Pierre Roux, Piotr Dollar, Polina Zvyagina, Prashant Ratanchandani, Pritish Yuvraj, Qian Liang, Rachad Alao, Rachel Rodriguez, Rafi Ayub, Raghotham Murthy, Raghu Nayani, Rahul Mitra, Rangaprabhu Parthasarathy, Raymond Li, Rebekkah Hogan, Robin Battey, Rocky Wang, Russ Howes, Ruty Rinott, Sachin Mehta, Sachin Siby, Sai Jayesh Bondu, Samyak Datta, Sara Chugh, Sara Hunt, Sargun Dhillon, Sasha Sidorov, Satadru Pan, Saurabh Mahajan, Saurabh Verma, Seiji Yamamoto, Sharadh Ramaswamy, Shaun Lindsay, Shaun Lindsay, Sheng Feng, Shenghao Lin, Shengxin Cindy Zha, Shishir Patil, Shiva Shankar, Shuqiang Zhang, Shuqiang Zhang, Sinong Wang, Sneha Agarwal, Soji Sajuyigbe, Soumith Chintala, Stephanie Max, Stephen